\pdfoutput=1

\documentclass[11pt]{article}

\usepackage[preprint]{acl}

\usepackage{times}
\usepackage{latexsym}
\usepackage{color,soul}

\usepackage[T1]{fontenc}

\usepackage[utf8]{inputenc}

\usepackage{microtype}

\usepackage{inconsolata}

\usepackage{graphicx}

\usepackage{booktabs}
\usepackage{enumitem}
\usepackage{amsmath}
\usepackage{algorithm}
\usepackage{algpseudocode}
\usepackage{makecell}
\usepackage{cleveref}

\title{Staircase Streaming for Low-Latency Multi-Agent Inference}

\author{\ Junlin Wang\textsuperscript{1}, Jue Wang \textsuperscript{2},  Zhen (Zach) Xu \textsuperscript{3}, Ben Athiwaratkun \textsuperscript{2} , \\ \textbf{Bhuwan Dhingra \textsuperscript{1}}, \textbf{Ce Zhang \textsuperscript{2,3}}, \textbf{James Zou \textsuperscript{2,4}} \\
\textsuperscript{1}Duke University \quad \textsuperscript{2}Together AI \quad\textsuperscript{3}University of Chicago \quad 
\textsuperscript{4}Stanford University }

\begin{document}
\maketitle
\begin{abstract}
Recent advances in large language models (LLMs) opened up new directions for leveraging the collective expertise of multiple LLMs. 
These methods, such as Mixture-of-Agents, typically employ additional inference steps to generate intermediate outputs, which are then used to produce the final response. 
While multi-agent inference can enhance response quality, it can significantly increase the time to first token (TTFT), posing a challenge for latency-sensitive applications and hurting user experience.
To address this issue, we propose \textit{staircase streaming} for low-latency multi-agent inference. 
Instead of waiting for the complete intermediate outputs from previous steps,
we begin generating the final response as soon as we receive partial outputs from these steps.
Experimental results demonstrate that staircase streaming reduces TTFT by up to 93\% while maintaining response quality.
\end{abstract}

\section{Introduction}

Large language models (LLMs) \citep{opt,palm,llama,gemini,gpt3,gpt4}
have significantly advanced the field of natural language processing, showing remarkable capabilities across a wide range of tasks and domains.
With the proliferation of LLMs and their impressive achievements, a new line of research has emerged: 
\textit{multi-agent inference} \citep{debate,mad,reconcile,moa}.
It aims to harness the collective strengths of multiple LLMs to produce more reliable, consistent, and high-quality responses, 
matching or even surpassing state-of-the-art performance.

While multi-agent inference techniques can significantly enhance response quality by combining the outputs of multiple LLMs, 
they often incorporate additional inference steps.
For example, Multi-Agent Debate (MAD) \citep{mad} uses several models to generate answers independently and then debate their answers;
Mixture-of-Agents (MoA) \citep{moa} employs proposers to generate initial responses and an aggregator to synthesize a higher-quality response from them.
They both require to wait for all additional generations to finish before producing the final output,
leading to an increased time to first token (TTFT).
This latency is particularly problematic for real-time applications such as chatbots,
where low latency is crucial for user experience. 
This naturally raises the question: \textit{How can we leverage the strengths of multi-agent inference while minimizing such latency?}


To address this question, we propose \textit{staircase streaming}, a novel approach for low-latency multi-agent inference. 
Instead of waiting for complete results from all the agents in the previous step, our approach begins streaming tokens once the first chunk of tokens becomes available from the preceding step.
By leveraging partial results and overlapping the generation processes, staircase streaming effectively reduces latency while maintaining the benefits of multi-agent inference.

The intuition behind staircase streaming is based on two key observations:
(1) \textit{Effective partial prompting}: \citealt{jiang2023llmlingua} demonstrate that even partial tokens within a prompt can effectively instruct LLM.
    Similarly, we found partial generation results can provide valuable, diverse viewpoints for other LLMs.
(2) \textit{Diminishing returns in subsequent outputs}: The most crucial information in a model's response is often front-loaded. 
    The first portion of generation typically contains the main ideas, while subsequent parts generally add details, examples, or refinements to these core concepts.


\begin{figure*}
    \centering
    \includegraphics[width=\linewidth]{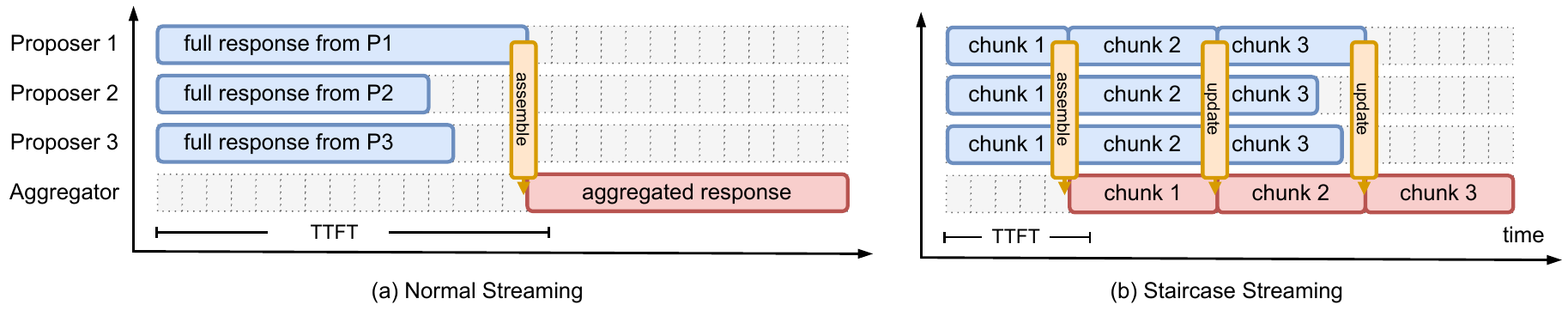}
    \caption{Comparison of normal streaming and staircase streaming using an MoA with 3 proposers and 1 aggregator. 
        (a) In normal streaming, each LLM generates a full response before proceeding to the next step, leading to a longer TTFT. 
        (b) Staircase streaming reduces TTFT by initiating the next step once the first chunks of proposed responses are available, enabling parallel processing between the proposers and the aggregator.}
    \label{fig:timeline}
\end{figure*}


\Cref{fig:timeline} illustrates an example of staircase streaming applied to MoA. 
In the normal case (a), the multi-agent sytem is bottlenecked by the proposer with the longest generation time, resulting in significant latency. 
With staircase streaming (b), the aggregator begins generating output as soon as a chunk of tokens is available from each proposer.
Once the aggregator generates a chunk of tokens, it waits for the next chunk from the proposers, updates its prompt with these newly received tokens, and continues generating. 

The evaluation results on the Arena-Hard and AlpacaEval benchmarks \citep{li2024crowdsourced,alpaca-eval}
demonstrate that our staircase streaming can be effectively adapted to multi-agent inference methods, 
reducing latency up to 93\% while maintaining the response quality.



\section{Methodology}

In this section, we first briefly overview token streaming. We then introduce staircase streaming, our approach that minimizes latency while leveraging multi-agent inference strengths. Additionally, we present a prefix caching optimized variant to further enhance efficiency.

\subsection{Token Streaming}

Token streaming allows autoregressive LMs to generate and return text incrementally, 
providing users with an early indication of content quality and enhancing the user experience. 
TTFT, measuring the latency between initiating a request and generating the first token, is a crucial metric for it.

However, multi-agent approaches often struggle with token streaming due to dependencies between multiple models. 
The first token becomes available after all preceding steps complete and the last step's prefill finishes.
Depending on the generation length, this process can incur substantial latency, sometimes spanning dozens of seconds.

\subsection{Staircase Streaming}



Staircase streaming is proposed to enable low-latency multi-agent inference by streaming tokens between models as soon as partial results are available, rather than waiting for complete outputs.
The key idea is to break the strict sequential dependencies between models,
allowing a pipelined execution pattern where models process partial results immediately.

\begin{algorithm}[t]
\scriptsize
\caption{Staircase Streaming for MoA}
\label{alg:ss}
\begin{algorithmic}[1]
\State \textbf{Input:} User prompt $Q$; Proposers \(P_1, ..., P_N\); Aggregator \(A\); Chunk sizes \(C_{\text{P}, j}\) and \(C_{\text{A}, j}\) ($j=1,2,...$)

\vspace{0.4em}
\State \textbf{(1) THREAD PROPOSER ($i=1,...,N$)}:
\State \quad Initialize \(j \leftarrow 0\)
\State \quad \textbf{while} not end of response
\State \quad \quad $j \leftarrow j + 1$
\State \quad \quad $R_{i,j}\leftarrow P_i(Q + \cup_{k=1}^{j-1} R_{i,k}) $ \Comment{Generate \(C_{\text{P}, j}\) tokens}
\State \quad \quad Send chunk $R_{i,j}$ to aggregator $A$
\State \quad \textbf{end while}

\vspace{0.4em}
\State \textbf{(2) THREAD AGGREGATOR}:
\State \quad Initialize \(j \leftarrow 0\)
\State \quad \textbf{while} not end of response
    \State \quad\quad $j \leftarrow j + 1$
    \State \quad\quad Receive chunks $R_{1,j}, ..., R_{N,j}$ from all proposers or total number of proposers minus redundancy if redundancy is not 0.
    \State \quad\quad Update prompt $Q_{\text{A},j}$ for the aggregator with proposed responses so far $[\cup_{k=1}^{j} R_{1,j}, ..., \cup_{k=1}^{j} R_{N,j}]$ with template in \Cref{tab:template-moa-ss}.
    \State \quad\quad $S_j \leftarrow A(Q_{\text{A},j} + \cup_{k=1}^{j-1} S_{k}) $ \Comment{Generate \(C_{\text{A}, j}\) tokens}
    \State \quad\quad (Streaming $S_j$ to the user)
\State \quad \textbf{end while}

\vspace{0.4em}
\State \textbf{Output: $\cup_{k=1}^{j} S_{k}$}
\end{algorithmic}
\end{algorithm}

MoA is a typical multi-agent inference system where multiple proposer models generate initial responses based on a user prompt.
These responses are then synthesized by an aggregator model to produce the final output.
\Cref{alg:ss} illustrates an adaptation of staircase streaming for MoA.
The proposers and aggregator run in separate threads. 
Each proposer generates a chunk of tokens based on the user prompt and previously generated chunks, and then immediately sends the new chunk to the aggregator.
The aggregator, after receiving chunks from proposers, updates its prompt by concatenating the new chunks.
Then it generates the next chunk of the final output based on the updated prompt, which is streamed to the user.
Meanwhile, the proposers process the next chunk of text in parallel.
For multi-layer MoA systems, the same staircase streaming approach can be applied to each intermediate LLM, creating a nested staircase streaming pipeline.

The chunk sizes \(C_{\text{P},j}\) and \(C_{\text{A},j}\) control the granularity of the streaming process and can be adjusted to balance latency and computational efficiency. 
Smaller chunk sizes lead to more frequent updates but may require more prefill operations, while larger chunk sizes can increase TTFT. 
To achieve a better trade-off, we use smaller chunk sizes at the beginning and gradually increase them as the generation continues.


Following the similar principle, staircase streaming can be applied to general multi-agent systems by having each agent generate and stream chunks of tokens in parallel. 
These chunks are passed to the next agent or aggregator, which updates its prompt and continues the process. 

\paragraph{Reducing TTFT}
In staircase streaming, TTFT is reduced since the aggregator starts generation with the first chunk \(R_{i,1}\) from each proposer rather than waiting for the complete responses \(R_i\). 
TTFT of MoA and staircase version can be represented as:

\vspace{-1em}
{\small
\begin{align}
    \text{TTFT}_{\text{normal}} &= \max_{1 \le i \le N} \left( \sum_{j=1}^{\text{eos}} T_{R_{i, j}} \right) + T_{\text{prefill}} \\
   \text{TTFT}_{\text{staircase}} &= \max_{1 \le i \le N} \left( T_{R_{i,1}} \right) + T_{\text{prefill}} \label{eq:ttft}
\end{align}}
where \(T_{R_{i,j}}\) is the time taken by proposer \(i\) to generate the $j$-th chunk and \(T_{\text{prefill}}\) is the time taken by the aggregator to process the initial chunks.

\subsection{Prefix-Caching Optimized Staircase Streaming}



Prefix-caching enhances the efficiency of transformer models by reusing previously computed key-value pairs for identical prompt prefixes \citep{zheng2023efficiently}. 
In staircase streaming, we can leverage this technique by appending new token chunks to the end of the prompt, keeping the prefix unchanged. 
This approach, whose prompt template is presented in \Cref{tab:template-moa-poss}, 
allows seamless updating of the staircase streaming prompt without requiring a full prefill computation.
While this method may fragment responses from the same proposer, it can save more compute.

\section{Evaluation}

\subsection{Setup}

\paragraph{Benchmarks} We evaluate on Arena-Hard \citep{li2024crowdsourced} and AlpacaEval 2.0 \citep{alpaca-eval}.
For efficiency metrics, we assess the time to first token (TTFT) and the end-to-end tokens per second (TPS)\footnote{
TPS = The number tokens generated by the aggregator $\div$ the time elapsed between sending the request and receiving the last token.
} on a subset of Arena-Hard and AlpacaEval with 64 samples (a total of 128 samples).

\paragraph{Methods}
We incorporate staircase streaming to Mixture-of-Agents (MoA) and Multi-Agent Debate (MAD).
We use Gemma-2-9B-IT \citep{team2024gemma}, LLaMA-3.1-8B-Instruct \citep{dubey2024llama}, Mistral-7B-Instruct-v0.3 \citep{jiang2023mistral}, and Qwen-1.5-7B-Chat \citep{bai2023qwen}; and use Gemma-2-9B-IT as the aggregator.
We set the chunk size to 8 for the first chunk, increasing to 128 for the second chunk, and capping at 256 for proposers and 128 for the aggregator for later chunks. 
Benchmarks were conducted using the Together Inference API.\footnote{\url{https://api.together.ai/playground/chat}} 
To minimize the effect of queuing times associated with the serverless API, we use redundancy of 2 -- we skip the slowest 2 models -- for the first chunk.
To reduce the variance caused by network latency and server load, 
we run the aggregator model with vLLM on 1 H100 GPU for TTFT and TPS metrics.
Detailed setup can be found in \Cref{sec:detailed-setups}.

\subsection{Results}

\begin{table}
    \centering
    \small
    \setlength{\tabcolsep}{1.5pt} 
    \begin{tabular}{@{}lcccc@{}}
    \toprule
         &  ArenaHard&  AlpacaEval&  TTFT & TPS \\
         &  Win Rate &  LC Win &  second& tokens/s \\
\midrule
 Gemma-2-9B-IT & 40.6& 48.5& 0.06$_{\pm0.01}$&69.4$_{\pm0.4}$\\
    \midrule
         MAD&  50.8&  55.8&  6.70$_{\pm2.3}$ & 35.9$_{\pm4.4}$\\
        \,\, + staircase&  45.9&  55.2&  0.45$_{\pm0.0}$& 44.7$_{\pm6.6}$\\
    \midrule
         MoA&  47.5&  56.8&  10.6$_{\pm3.4}$& 28.3$_{\pm8.9}$\\
\,\, + staircase& 48.3& 55.1& 0.47$_{\pm0.1}$&43.4$_{\pm6.5}$\\
            \makecell[l]{\,\, + staircase \& \\ \,\,\,\, prefix-cache }
                    & 46.9& 53.0& 0.45$_{\pm0.1}$&45.0$_{\pm7.0}$\\
    \bottomrule
    \end{tabular}
    \caption{Results of different inference methods.
    }
    \label{tab:results}
\end{table}

\paragraph{Inference Efficiency and Chat Benchmarks}
\Cref{tab:results} presents the evaluation results.
Our approach significantly reduces TTFT by up to 93\% and increases TPS by up to 1.6x, 
enhancing the responsiveness while preserving the response quality of multi-agent inference.
Staircase streaming is effective across MoA and MAD, showing its versatility.
While the prefix-cache optimized variant slightly reduces the win rate, it further improves the efficiency. For tasks with critical information at the end (reasoning tasks), our method maintains performance due to larger proposer chunks allowing better aggregator capacity. 

\paragraph{Reasoning Capability}
\Cref{tab:cot_results} for reasoning task results. 
MoA with staircase streaming improve over the best proposers,
while maintaining an acceptable drop compared to the original MoA.

\begin{figure}
    \centering
    \includegraphics[width=0.85\linewidth]{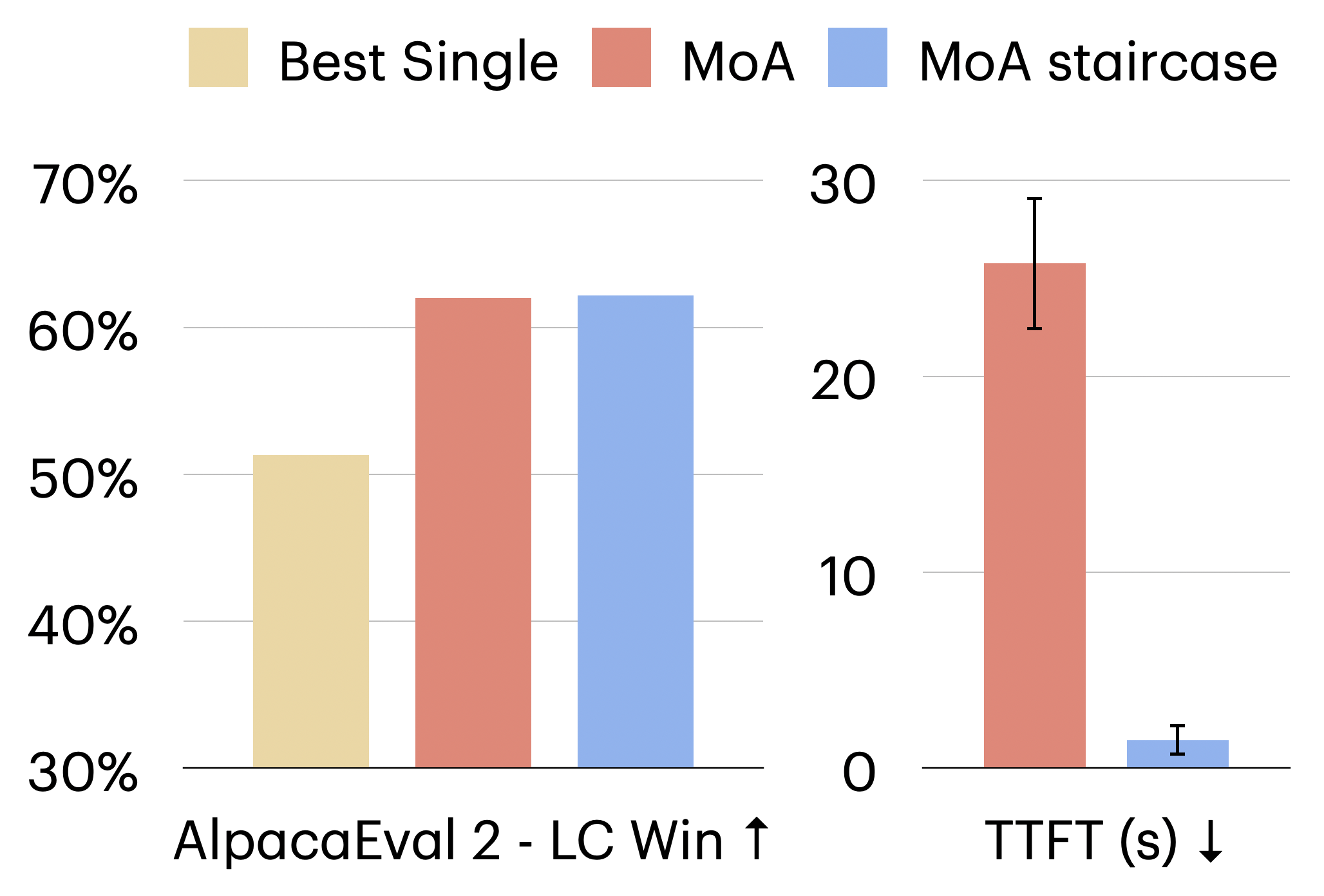}
    \caption{Results on Larger LLMs. 
             The `Best Single' model is WizardLM 8x22B. 
             For MoA, the proposers include Qwen1.5-72B-Chat, Qwen1.5-110B-Chat, Wizard 8x22B, Mixtral-8x22B-Instruct-v0.1, and Llama-3-70B-Instruct. 
             The aggregator is Qwen1.5-110B-Chat.
             TTFT was evaluated using the Together serverless endpoint, so the results may vary depending on server load. 
    }
    \label{fig:large-models}
\end{figure}

\paragraph{Scaling up Model Size}
\Cref{fig:large-models} shows AlpacaEval results for larger models. The 'Best Single' model achieved a 51\% LC win rate. 
In contrast, MoA with 6 proposers achieved 62\%, showing the benefits of leveraging the expertise from multiple LLMs.
The staircase version achieved a similar LC win rate while maintaining reasonable TTFT, showing our method effectively scales to larger models, achieving good performance with lower latency. 

\begin{figure}
    \centering
    \includegraphics[width=0.9\linewidth]{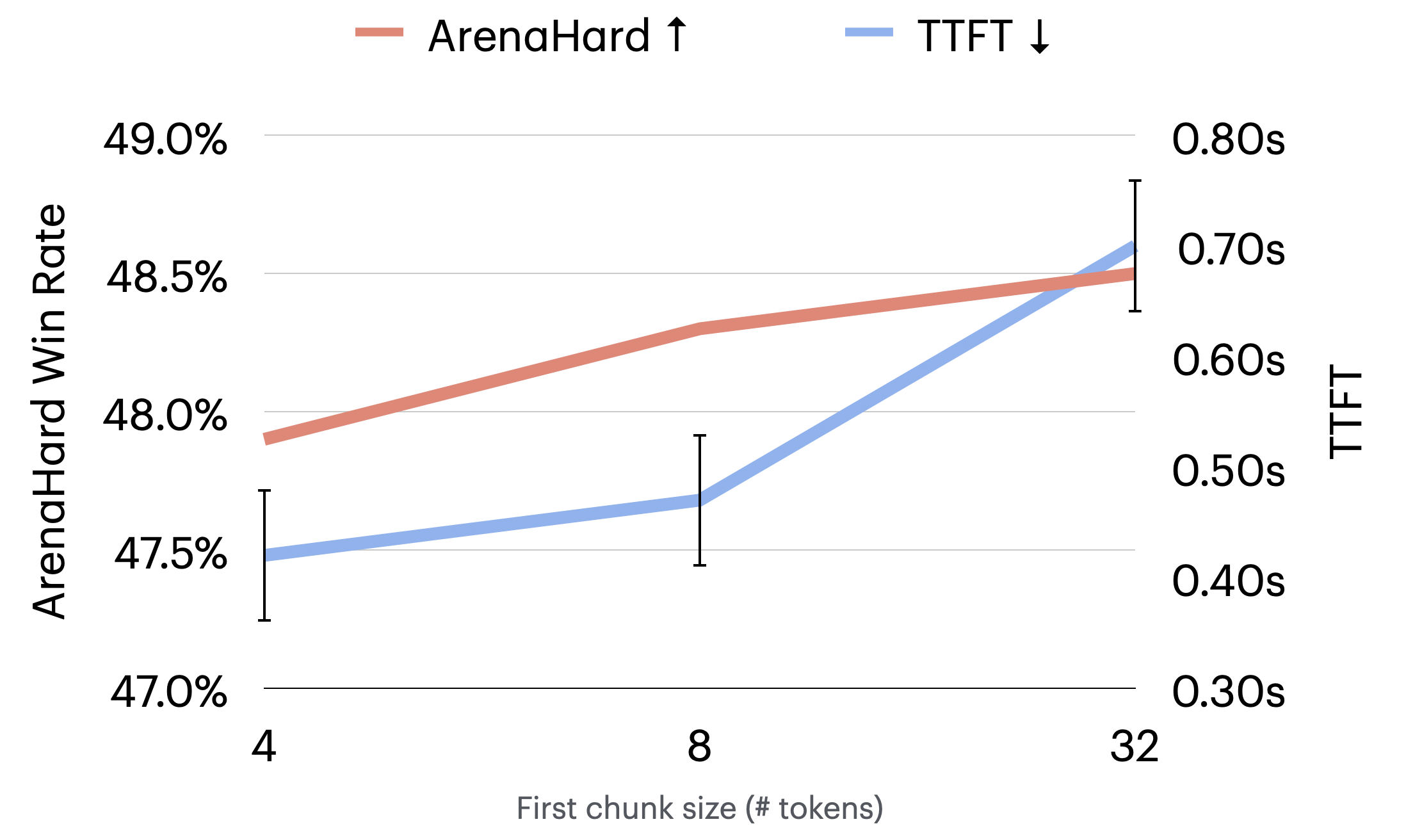}
    \caption{The impact of first chunk size on ArenaHard win rate and TTFT.}
    \label{fig:chunk-size}
\end{figure}

\paragraph{Effect of Chunk Sizes}
The first chunk size is a crucial hyperparameter for optimizing TTFT, as indicated in \cref{eq:ttft}.
\Cref{fig:chunk-size} presents the impact of varying the first chunk size on response quality and TTFT. 
As the chunk size increases from 4 to 32 tokens, the win rate improves, while TTFT initially decreases but then increases for larger chunk sizes.
This presents the trade-off between quality and latency, with a chunk size of 8 as the sweet point.


\paragraph{Case Study}
\Cref{tab:case,tab:case2} present case studies demonstrating how staircase streaming effectively leverages the proposed chunks while maintaining coherence in context.

\section{Related Work}
\label{section:related_work}

\paragraph{Multi-Agent Inference}
To mitigate the computational costs associated with multi-LLM inference,
previous studies have explored training a \textit{router} that predicts the best-performing model from a fixed set of LLMs for a given input \citep{wang2024fusing,shnitzer2024large,lu2023routing,chen2023frugalgpt}. 
Another line of work focuses on collaborative inference between multiple LLMs \citep{chan2023chateval,debate,mad,reconcile,moa}, where models work together to generate more reliable, consistent, and high-quality responses.

\paragraph{Efficient Inference}
Recent LLM inference optimizations enhance efficiency and resource use. 
PagedAttention~\citep{vllm} is proposed to manage non-contiguous memory blocks efficiently.
To further address the memory consumption of key-value caches, recent research proposes using quantization and sparsification techniques \citep{kvquant,h2o,streamingllm}.
Speculative decoding~\citep{speculative,speculativesampling} speeds up inference by using extra small models to generate multiple token guesses and validate them in parallel.
To the best of our knowledge, there has been limited work for the inference efficiency of multi-agent systems.

\section{Conclusion}

In this paper, we introduced staircase streaming, an effective approach to reduce TTFT in multi-agent systems by streaming tokens incrementally between models. 
The key innovation is to break the strict sequential dependencies inherent in existing multi-agent systems, 
enabling a pipelined execution pattern that leverages partial results to minimize idle waiting time.
Our results showed that this approach not only maintains response quality but also significantly reduces TTFT by up to 93\%.

\section{Limitation}
Our method requires modification to the original multi-agent algorithm. 
While these changes are generally minimal, they may impact performance.
So necessary evaluation is needed before adapting staircase streaming to other multi-agent methods to ensure the quality does not change significantly.

Additionally, our work focuses primarily on TTFT as it is crucial for chat applications. However, we think chunk-by-chunk latency can be an interesting direction for future work.




\section{Ethical Considerations}
 
Although multi-agent inference generally improves the harmlessness of model outputs, as shown in the Mixture-of-Agent paper \citep{moa}, there are still potential risks that the final answer may retain biases or harmful content from the intermediate outputs. 
This underscores the importance of diligent oversight and ethical guidelines in the development and deployment of such systems.

\bibliography{custom}

\begin{thebibliography}{41}
\providecommand{\natexlab}[1]{#1}

\bibitem[{Aminabadi et~al.(2022)Aminabadi, Rajbhandari, Awan, Li, Li, Zheng, Ruwase, Smith, Zhang, Rasley, and He}]{deepspeedinference}
Reza~Yazdani Aminabadi, Samyam Rajbhandari, Ammar~Ahmad Awan, Cheng Li, Du~Li, Elton Zheng, Olatunji Ruwase, Shaden Smith, Minjia Zhang, Jeff Rasley, and Yuxiong He. 2022.
\newblock Deepspeed-inference: enabling efficient inference of transformer models at unprecedented scale.
\newblock In \emph{Proceedings of the International Conference on High Performance Computing, Networking, Storage and Analysis}, SC '22. IEEE Press.

\bibitem[{Bai et~al.(2023)Bai, Bai, Chu, Cui, Dang, Deng, Fan, Ge, Han, Huang, Hui, Ji, Li, Lin, Lin, Liu, Liu, Lu, Lu, Ma, Men, Ren, Ren, Tan, Tan, Tu, Wang, Wang, Wang, Wu, Xu, Xu, Yang, Yang, Yang, Yang, Yao, Yu, Yuan, Yuan, Zhang, Zhang, Zhang, Zhang, Zhou, Zhou, Zhou, and Zhu}]{bai2023qwen}
Jinze Bai, Shuai Bai, Yunfei Chu, Zeyu Cui, Kai Dang, Xiaodong Deng, Yang Fan, Wenbin Ge, Yu~Han, Fei Huang, Binyuan Hui, Luo Ji, Mei Li, Junyang Lin, Runji Lin, Dayiheng Liu, Gao Liu, Chengqiang Lu, Keming Lu, Jianxin Ma, Rui Men, Xingzhang Ren, Xuancheng Ren, Chuanqi Tan, Sinan Tan, Jianhong Tu, Peng Wang, Shijie Wang, Wei Wang, Shengguang Wu, Benfeng Xu, Jin Xu, An~Yang, Hao Yang, Jian Yang, Shusheng Yang, Yang Yao, Bowen Yu, Hongyi Yuan, Zheng Yuan, Jianwei Zhang, Xingxuan Zhang, Yichang Zhang, Zhenru Zhang, Chang Zhou, Jingren Zhou, Xiaohuan Zhou, and Tianhang Zhu. 2023.
\newblock Qwen technical report.
\newblock \emph{arXiv preprint arXiv: 2309.16609}.

\bibitem[{Brown et~al.(2020)Brown, Mann, Ryder, Subbiah, Kaplan, Dhariwal, Neelakantan, Shyam, Sastry, Askell et~al.}]{gpt3}
Tom Brown, Benjamin Mann, Nick Ryder, Melanie Subbiah, Jared~D Kaplan, Prafulla Dhariwal, Arvind Neelakantan, Pranav Shyam, Girish Sastry, Amanda Askell, et~al. 2020.
\newblock Language models are few-shot learners.
\newblock \emph{Advances in neural information processing systems}, 33:1877--1901.

\bibitem[{Chan et~al.(2023)Chan, Chen, Su, Yu, Xue, Zhang, Fu, and Liu}]{chan2023chateval}
Chi-Min Chan, Weize Chen, Yusheng Su, Jianxuan Yu, Wei Xue, Shanghang Zhang, Jie Fu, and Zhiyuan Liu. 2023.
\newblock Chateval: Towards better llm-based evaluators through multi-agent debate.
\newblock \emph{arXiv preprint arXiv:2308.07201}.

\bibitem[{Chen et~al.(2023{\natexlab{a}})Chen, Borgeaud, Irving, Lespiau, Sifre, and Jumper}]{speculativesampling}
Charlie Chen, Sebastian Borgeaud, Geoffrey Irving, Jean-Baptiste Lespiau, L.~Sifre, and John~M. Jumper. 2023{\natexlab{a}}.
\newblock Accelerating large language model decoding with speculative sampling.
\newblock \emph{ArXiv}, abs/2302.01318.

\bibitem[{Chen et~al.(2023{\natexlab{b}})Chen, Saha, and Bansal}]{reconcile}
Justin Chih-Yao Chen, Swarnadeep Saha, and Mohit Bansal. 2023{\natexlab{b}}.
\newblock Reconcile: Round-table conference improves reasoning via consensus among diverse llms.
\newblock \emph{arXiv preprint arXiv:2309.13007}.

\bibitem[{Chen et~al.(2023{\natexlab{c}})Chen, Zaharia, and Zou}]{chen2023frugalgpt}
Lingjiao Chen, Matei Zaharia, and James Zou. 2023{\natexlab{c}}.
\newblock Frugalgpt: How to use large language models while reducing cost and improving performance.
\newblock \emph{arXiv preprint arXiv:2305.05176}.

\bibitem[{Chowdhery et~al.(2022)Chowdhery, Narang, Devlin, Bosma, Mishra, Roberts, Barham, Chung, Sutton, Gehrmann et~al.}]{palm}
Aakanksha Chowdhery, Sharan Narang, Jacob Devlin, Maarten Bosma, Gaurav Mishra, Adam Roberts, Paul Barham, Hyung~Won Chung, Charles Sutton, Sebastian Gehrmann, et~al. 2022.
\newblock Palm: Scaling language modeling with pathways.
\newblock \emph{arXiv preprint arXiv:2204.02311}.

\bibitem[{Cobbe et~al.(2021)Cobbe, Kosaraju, Bavarian, Chen, Jun, Kaiser, Plappert, Tworek, Hilton, Nakano et~al.}]{gsm8k}
Karl Cobbe, Vineet Kosaraju, Mohammad Bavarian, Mark Chen, Heewoo Jun, Lukasz Kaiser, Matthias Plappert, Jerry Tworek, Jacob Hilton, Reiichiro Nakano, et~al. 2021.
\newblock Training verifiers to solve math word problems.
\newblock \emph{arXiv preprint arXiv:2110.14168}.

\bibitem[{Dao et~al.(2022)Dao, Fu, Ermon, Rudra, and R'e}]{flashattn}
Tri Dao, Daniel~Y. Fu, Stefano Ermon, Atri Rudra, and Christopher R'e. 2022.
\newblock Flashattention: Fast and memory-efficient exact attention with io-awareness.
\newblock \emph{ArXiv}, abs/2205.14135.

\bibitem[{Dao et~al.(2023)Dao, Haziza, Massa, and Sizov}]{flashdecoding}
Tri Dao, Daniel Haziza, Francisco Massa, and Grigory Sizov. 2023.
\newblock Flash-decoding for long-context inference.

\bibitem[{Du et~al.(2023)Du, Li, Torralba, Tenenbaum, and Mordatch}]{debate}
Yilun Du, Shuang Li, Antonio Torralba, Joshua~B Tenenbaum, and Igor Mordatch. 2023.
\newblock Improving factuality and reasoning in language models through multiagent debate.
\newblock \emph{arXiv preprint arXiv:2305.14325}.

\bibitem[{Dubey et~al.(2024)Dubey, Jauhri, Pandey, Kadian, Al-Dahle, Letman, Mathur, Schelten, Yang, Fan et~al.}]{dubey2024llama}
Abhimanyu Dubey, Abhinav Jauhri, Abhinav Pandey, Abhishek Kadian, Ahmad Al-Dahle, Aiesha Letman, Akhil Mathur, Alan Schelten, Amy Yang, Angela Fan, et~al. 2024.
\newblock The llama 3 herd of models.
\newblock \emph{arXiv preprint arXiv:2407.21783}.

\bibitem[{Dubois et~al.(2024)Dubois, Galambosi, Liang, and Hashimoto}]{alpaca-eval}
Yann Dubois, Bal{\'a}zs Galambosi, Percy Liang, and Tatsunori~B Hashimoto. 2024.
\newblock Length-controlled alpacaeval: A simple way to debias automatic evaluators.
\newblock \emph{arXiv preprint arXiv:2404.04475}.

\bibitem[{Hendrycks et~al.(2021)Hendrycks, Burns, Kadavath, Arora, Basart, Tang, Song, and Steinhardt}]{hendrycksmath2021}
Dan Hendrycks, Collin Burns, Saurav Kadavath, Akul Arora, Steven Basart, Eric Tang, Dawn Song, and Jacob Steinhardt. 2021.
\newblock Measuring mathematical problem solving with the math dataset.
\newblock \emph{NeurIPS}.

\bibitem[{Hooper et~al.(2024)Hooper, Kim, Mohammadzadeh, Mahoney, Shao, Keutzer, and Gholami}]{kvquant}
Coleman Hooper, Sehoon Kim, Hiva Mohammadzadeh, Michael~W. Mahoney, Yakun~Sophia Shao, Kurt Keutzer, and Amir Gholami. 2024.
\newblock Kvquant: Towards 10 million context length llm inference with kv cache quantization.
\newblock \emph{ArXiv}, abs/2401.18079.

\bibitem[{Huang et~al.(2024)Huang, Feng, Li, Xiang, Wang, Qin, and Liu}]{huang2024enabling}
Yichong Huang, Xiaocheng Feng, Baohang Li, Yang Xiang, Hui Wang, Bing Qin, and Ting Liu. 2024.
\newblock Enabling ensemble learning for heterogeneous large language models with deep parallel collaboration.
\newblock \emph{arXiv preprint arXiv:2404.12715}.

\bibitem[{Jiang et~al.(2023{\natexlab{a}})Jiang, Sablayrolles, Mensch, Bamford, Chaplot, de~las Casas, Bressand, Lengyel, Lample, Saulnier, Lavaud, Lachaux, Stock, Scao, Lavril, Wang, Lacroix, and Sayed}]{jiang2023mistral}
Albert~Q. Jiang, Alexandre Sablayrolles, Arthur Mensch, Chris Bamford, Devendra~Singh Chaplot, Diego de~las Casas, Florian Bressand, Gianna Lengyel, Guillaume Lample, Lucile Saulnier, Lélio~Renard Lavaud, Marie-Anne Lachaux, Pierre Stock, Teven~Le Scao, Thibaut Lavril, Thomas Wang, Timothée Lacroix, and William~El Sayed. 2023{\natexlab{a}}.
\newblock \href {https://arxiv.org/abs/2310.06825} {Mistral 7b}.
\newblock \emph{Preprint}, arXiv:2310.06825.

\bibitem[{Jiang et~al.(2023{\natexlab{b}})Jiang, Ren, and Lin}]{jiang-etal-2023-llm}
Dongfu Jiang, Xiang Ren, and Bill~Yuchen Lin. 2023{\natexlab{b}}.
\newblock \href {https://doi.org/10.18653/v1/2023.acl-long.792} {{LLM}-blender: Ensembling large language models with pairwise ranking and generative fusion}.
\newblock In \emph{Proceedings of the 61st Annual Meeting of the Association for Computational Linguistics (Volume 1: Long Papers)}, pages 14165--14178, Toronto, Canada. Association for Computational Linguistics.

\bibitem[{Jiang et~al.(2023{\natexlab{c}})Jiang, Wu, Lin, Yang, and Qiu}]{jiang2023llmlingua}
Huiqiang Jiang, Qianhui Wu, Chin-Yew Lin, Yuqing Yang, and Lili Qiu. 2023{\natexlab{c}}.
\newblock Llmlingua: Compressing prompts for accelerated inference of large language models.
\newblock \emph{arXiv preprint arXiv:2310.05736}.

\bibitem[{Kim et~al.(2023)Kim, Mangalam, Moon, Malik, Mahoney, Gholami, and Keutzer}]{bild}
Sehoon Kim, Karttikeya Mangalam, Suhong Moon, Jitendra Malik, Michael~W. Mahoney, Amir Gholami, and Kurt Keutzer. 2023.
\newblock Speculative decoding with big little decoder.
\newblock In \emph{Neural Information Processing Systems}.

\bibitem[{Kwon et~al.(2023)Kwon, Li, Zhuang, Sheng, Zheng, Yu, Gonzalez, Zhang, and Stoica}]{vllm}
Woosuk Kwon, Zhuohan Li, Siyuan Zhuang, Ying Sheng, Lianmin Zheng, Cody~Hao Yu, Joseph~E. Gonzalez, Hao Zhang, and Ion Stoica. 2023.
\newblock Efficient memory management for large language model serving with pagedattention.
\newblock In \emph{Proceedings of the ACM SIGOPS 29th Symposium on Operating Systems Principles}.

\bibitem[{Leviathan et~al.(2022)Leviathan, Kalman, and Matias}]{speculative}
Yaniv Leviathan, Matan Kalman, and Yossi Matias. 2022.
\newblock Fast inference from transformers via speculative decoding.
\newblock In \emph{International Conference on Machine Learning}.

\bibitem[{Li et~al.(2024)Li, Chiang, Frick, Dunlap, Wu, Zhu, Gonzalez, and Stoica}]{li2024crowdsourced}
Tianle Li, Wei-Lin Chiang, Evan Frick, Lisa Dunlap, Tianhao Wu, Banghua Zhu, Joseph~E Gonzalez, and Ion Stoica. 2024.
\newblock From crowdsourced data to high-quality benchmarks: Arena-hard and benchbuilder pipeline.
\newblock \emph{arXiv preprint arXiv:2406.11939}.

\bibitem[{Liang et~al.(2023)Liang, He, Jiao, Wang, Wang, Wang, Yang, Tu, and Shi}]{mad}
Tian Liang, Zhiwei He, Wenxiang Jiao, Xing Wang, Yan Wang, Rui Wang, Yujiu Yang, Zhaopeng Tu, and Shuming Shi. 2023.
\newblock Encouraging divergent thinking in large language models through multi-agent debate.
\newblock \emph{arXiv preprint arXiv:2305.19118}.

\bibitem[{Lu et~al.(2023)Lu, Yuan, Lin, Lin, Yuan, Zhou, and Zhou}]{lu2023routing}
Keming Lu, Hongyi Yuan, Runji Lin, Junyang Lin, Zheng Yuan, Chang Zhou, and Jingren Zhou. 2023.
\newblock \href {https://arxiv.org/abs/2311.08692} {Routing to the expert: Efficient reward-guided ensemble of large language models}.
\newblock \emph{Preprint}, arXiv:2311.08692.

\bibitem[{OpenAI(2023)}]{gpt4}
OpenAI. 2023.
\newblock \href {https://arxiv.org/abs/2303.08774} {Gpt-4 technical report}.
\newblock \emph{Preprint}, arXiv:2303.08774.

\bibitem[{Shazeer et~al.(2017)Shazeer, Mirhoseini, Maziarz, Davis, Le, Hinton, and Dean}]{moe}
Noam Shazeer, Azalia Mirhoseini, Krzysztof Maziarz, Andy Davis, Quoc Le, Geoffrey Hinton, and Jeff Dean. 2017.
\newblock Outrageously large neural networks: The sparsely-gated mixture-of-experts layer.
\newblock \emph{arXiv preprint arXiv:1701.06538}.

\bibitem[{Sheng et~al.(2023)Sheng, Zheng, Yuan, Li, Ryabinin, Chen, Liang, R\'{e}, Stoica, and Zhang}]{flexgen}
Ying Sheng, Lianmin Zheng, Binhang Yuan, Zhuohan Li, Max Ryabinin, Beidi Chen, Percy Liang, Christopher R\'{e}, Ion Stoica, and Ce~Zhang. 2023.
\newblock Flexgen: high-throughput generative inference of large language models with a single gpu.
\newblock In \emph{Proceedings of the 40th International Conference on Machine Learning}, ICML'23. JMLR.org.

\bibitem[{Shnitzer et~al.(2024)Shnitzer, Ou, Silva, Soule, Sun, Solomon, Thompson, and Yurochkin}]{shnitzer2024large}
Tal Shnitzer, Anthony Ou, M{\'\i}rian Silva, Kate Soule, Yuekai Sun, Justin Solomon, Neil Thompson, and Mikhail Yurochkin. 2024.
\newblock \href {https://openreview.net/forum?id=LyNsMNNLjY} {Large language model routing with benchmark datasets}.

\bibitem[{Spector and Re(2023)}]{stagedspeculative}
Benjamin Spector and Christal Re. 2023.
\newblock Accelerating llm inference with staged speculative decoding.
\newblock \emph{ArXiv}, abs/2308.04623.

\bibitem[{Team et~al.(2023)Team, Anil, Borgeaud, Wu, Alayrac, Yu, Soricut, Schalkwyk, Dai, Hauth et~al.}]{gemini}
Gemini Team, Rohan Anil, Sebastian Borgeaud, Yonghui Wu, Jean-Baptiste Alayrac, Jiahui Yu, Radu Soricut, Johan Schalkwyk, Andrew~M Dai, Anja Hauth, et~al. 2023.
\newblock Gemini: a family of highly capable multimodal models.
\newblock \emph{arXiv preprint arXiv:2312.11805}.

\bibitem[{Team et~al.(2024)Team, Riviere, Pathak, Sessa, Hardin, Bhupatiraju, Hussenot, Mesnard, Shahriari, Ram{\'e} et~al.}]{team2024gemma}
Gemma Team, Morgane Riviere, Shreya Pathak, Pier~Giuseppe Sessa, Cassidy Hardin, Surya Bhupatiraju, L{\'e}onard Hussenot, Thomas Mesnard, Bobak Shahriari, Alexandre Ram{\'e}, et~al. 2024.
\newblock Gemma 2: Improving open language models at a practical size.
\newblock \emph{arXiv preprint arXiv:2408.00118}.

\bibitem[{Touvron et~al.(2023)Touvron, Lavril, Izacard, Martinet, Lachaux, Lacroix, Rozi{\`e}re, Goyal, Hambro, Azhar et~al.}]{llama}
Hugo Touvron, Thibaut Lavril, Gautier Izacard, Xavier Martinet, Marie-Anne Lachaux, Timoth{\'e}e Lacroix, Baptiste Rozi{\`e}re, Naman Goyal, Eric Hambro, Faisal Azhar, et~al. 2023.
\newblock Llama: Open and efficient foundation language models.
\newblock \emph{arXiv preprint arXiv:2302.13971}.

\bibitem[{Wang et~al.(2024{\natexlab{a}})Wang, Polo, Sun, Kundu, Xing, and Yurochkin}]{wang2024fusing}
Hongyi Wang, Felipe~Maia Polo, Yuekai Sun, Souvik Kundu, Eric Xing, and Mikhail Yurochkin. 2024{\natexlab{a}}.
\newblock \href {https://openreview.net/forum?id=PhMrGCMIRL} {Fusing models with complementary expertise}.
\newblock In \emph{The Twelfth International Conference on Learning Representations}.

\bibitem[{Wang et~al.(2024{\natexlab{b}})Wang, Wang, Athiwaratkun, Zhang, and Zou}]{moa}
Junlin Wang, Jue Wang, Ben Athiwaratkun, Ce~Zhang, and James Zou. 2024{\natexlab{b}}.
\newblock Mixture-of-agents enhances large language model capabilities.
\newblock \emph{arXiv preprint arXiv:2406.04692}.

\bibitem[{Wang et~al.(2024{\natexlab{c}})Wang, Wang, Su, Tong, and Song}]{wang2024rethinking}
Qineng Wang, Zihao Wang, Ying Su, Hanghang Tong, and Yangqiu Song. 2024{\natexlab{c}}.
\newblock Rethinking the bounds of llm reasoning: Are multi-agent discussions the key?
\newblock \emph{arXiv preprint arXiv:2402.18272}.

\bibitem[{Xiao et~al.(2023)Xiao, Tian, Chen, Han, and Lewis}]{streamingllm}
Guangxuan Xiao, Yuandong Tian, Beidi Chen, Song Han, and Mike Lewis. 2023.
\newblock Efficient streaming language models with attention sinks.
\newblock \emph{ArXiv}, abs/2309.17453.

\bibitem[{Zhang et~al.(2022)Zhang, Roller, Goyal, Artetxe, Chen, Chen, Dewan, Diab, Li, Lin et~al.}]{opt}
Susan Zhang, Stephen Roller, Naman Goyal, Mikel Artetxe, Moya Chen, Shuohui Chen, Christopher Dewan, Mona Diab, Xian Li, Xi~Victoria Lin, et~al. 2022.
\newblock Opt: Open pre-trained transformer language models.
\newblock \emph{arXiv e-prints}, pages arXiv--2205.

\bibitem[{Zhang et~al.(2023)Zhang, Sheng, Zhou, Chen, Zheng, Cai, Song, Tian, R{\'e}, Barrett, Wang, and Chen}]{h2o}
Zhenyu~(Allen) Zhang, Ying Sheng, Tianyi Zhou, Tianlong Chen, Lianmin Zheng, Ruisi Cai, Zhao Song, Yuandong Tian, Christopher R{\'e}, Clark~W. Barrett, Zhangyang Wang, and Beidi Chen. 2023.
\newblock H2o: Heavy-hitter oracle for efficient generative inference of large language models.
\newblock \emph{ArXiv}, abs/2306.14048.

\bibitem[{Zheng et~al.(2023)Zheng, Yin, Xie, Huang, Sun, Yu, Cao, Kozyrakis, Stoica, Gonzalez et~al.}]{zheng2023efficiently}
Lianmin Zheng, Liangsheng Yin, Zhiqiang Xie, Jeff Huang, Chuyue Sun, Cody~Hao Yu, Shiyi Cao, Christos Kozyrakis, Ion Stoica, Joseph~E Gonzalez, et~al. 2023.
\newblock Efficiently programming large language models using sglang.
\newblock \emph{arXiv preprint arXiv:2312.07104}.

\end{thebibliography}

\appendix

\section{Detailed Setups}
\label{sec:detailed-setups}

In the code implementation, there are five hyperparameters that can control the behaviors of staircase streaming:
\textit{first\_chunk\_size}, \textit{second\_chunk\_size}, \textit{chunk\_size}, \textit{aggregator\_chunk\_size}, and \textit{redundancy}. 
When running staircase streaming, \(C_{\text{P},1}\) and \(C_{\text{A},1}\) would be set to \textit{first\_chunk\_size}.  \(C_{\text{P},2}\) and \(C_{\text{A},2}\) are set to \textit{second\_chunk\_size}. Then for the rest of the chunks, \(C_{\text{P},j}\) and \(C_{\text{A},j}\) are set to \textit{chunk\_size} and \textit{aggregator\_chunk\_size} respectively for \(j>2\). 

\textit{Redundancy} is introduced to control how many models' responses must be ready before the aggregator can start generating. 
For example, if there are five proposers and \textit{Redundancy}=2, then that means for the first chunk, it only needs responses from three models before the aggregator can generate the first chunk. 
This granular setup allows practitioners to have maximum control over the trade-offs between TTFT and performance. 

For our main results in Table \ref{tab:results}, we use \textit{first\_chunk\_size}=8, \textit{second\_chunk\_size}=128, \textit{chunk\_size}=256, \textit{aggregator\_chunk\_size}=128, and \textit{redundancy}=2. These choices is carefully chosen from the ablation study, and generalize across different setups.

The improved TTFT and streaming experience comes at a cost of increased prompt token consumption. To mitigate this, we proposed a prefix-caching optimized version that can save a significant number of the tokens if the API services or the local inference engine supports it. Currently, more and more inference engines \citep{zheng2023efficiently,vllm} and API services roll out the support prefix-caching.

\section{Benchmarks Used}

\paragraph{AlpacaEval 2} comprises 805 instructions that closely mirror real-world use cases. It implements length-controlled (LC) win rates to effectively neutralize length bias, a common confounding factor in language model evaluation. This metric has demonstrated remarkable alignment with human preferences, achieving a Spearman correlation of 0.98 with actual human evaluations. 

\paragraph{Arena-Hard-Auto} targets the evaluation of models on 500 challenging and demanding instructions submitted by real users in Chatbot Arena, thus maintaining a strong correlation with human preferences in complex scenarios. It contains difficult reasoning questions in the domain of math and coding.

\paragraph{MATH}\citep{hendrycksmath2021} contains 5,000 challenging math problems that can showcase the model's reasoning capability.

\section{Prompt Templates}
\label{sec:templates}

We present the prompt templates used in our evaluation.
For MoA experiments, we mainly followed the original paper by \cite{moa}. 
The staircase streaming templates are shown in \Cref{tab:template-moa-ss} and \Cref{tab:template-moa-poss}.
For MAD experiments, we adapted the templates to suit open-ended chat scenarios, as MAD is originally designed for tasks with short and deterministic answers, e.g.~classification. 
The staircase streaming templates is shown in \Cref{tab:template-mad-ss}.

\begin{table*}[t]
    \centering
    \small
    \caption{Prompt template of staircase streaming for MoA.}
    \begin{tabular}{@{}lp{0.9\linewidth}@{}}
    \toprule
    Role & Content \\
    \midrule
    system &
    You have been provided with a set of responses from various open-source models to the latest user query. Your task is to synthesize these responses into a single, high-quality response. It is crucial to critically evaluate the information provided in these responses, recognizing that some of it may be biased or incorrect. Your response should not simply replicate the given answers but should offer a refined, accurate, and comprehensive reply to the instruction. Ensure your response is well-structured, coherent, and adheres to the highest standards of accuracy and reliability.
    \\ \\
    & Responses from models:
    \\
    & 1. \{$\cup_{k=1}^{j} R_{1,k}$\} \\
    & 2. \{$\cup_{k=1}^{j} R_{2,k}$\} \\
    & ... \\
    & $N$. \{$\cup_{k=1}^{j} R_{N,k}$\} \\
    \midrule
    user & \{Q\} \\
    \bottomrule
    \end{tabular}
    \label{tab:template-moa-ss}
\end{table*}

\begin{table}
    \centering
    \small
    \setlength{\tabcolsep}{3pt} 
    \begin{tabular}{@{}lcc@{}}
    \toprule
    Model & MATH (\%) & GSM8k (\%) \\
    \midrule
    Gemma-2-9B-IT & 46.4 & 88.1 \\
    LLaMA-3.1-8B-Instruct & 44.1 & 71.6 \\
    Mistral-7B-Instruct-v0.3 & 11.5 & 42.1 \\
    Qwen-1.5-7B-Chat & 21.7 & 51.2 \\
    Qwen-2.5-7B-Instruct & 70.9 & 90.4 \\
    \midrule
    MoA & 47.0 & 78.1 \\
    \,\, + staircase & 46.6 & 82.4 \\
    \midrule
    MoA\_version2 & 73.5 & 92.7 \\
    \,\, + staircase & 72.2 & 91.7 \\
    \bottomrule
    \end{tabular}
    \caption{Results of different inference methods on the MATH \citep{hendrycksmath2021} and GSM8k \citep{gsm8k} datasets. \textit{MoA\_version2} replaces Qwen-1.5-7B-Chat with Qwen-2.5-7B-Instruct and uses Qwen-2.5-7B-Instruct as aggregator. While staircase results show slight degradation compared to MoA results on MATH, they improve over the best proposers for GSM8k.}
    \label{tab:cot_results}
\end{table}

\begin{table*}
    \centering
    \small
    \setlength{\tabcolsep}{2.5pt} 
    \begin{tabular}{@{}lcc@{}}
    \toprule
    Method & Arena-Hard & AlpacaEval2 (LC win rate) \\
    \midrule
    LLaMA-3.1-8B-Instruct & 28.0 & 26.06 \\
    Mistral-7B-Instruct-v0.3 & 16.3 & 19.88 \\
    Qwen-1.5-7B-Chat & 12.6 & 16.58 \\
    MoA w/ LLaMA-3.1-8B-Instruct Aggregator & 34.9 & 29.52 \\
    \,\, + staircase & 33.6 & 28.34 \\
    MoA w/ Mistral-7B-Instruct-v0.3 Aggregator & 22.1 & 26.93 \\
    \,\, + staircase & 23.8 & 25.39 \\
    MoA w/ Qwen-1.5-7B-Chat Aggregator & 20.7 & 25.54 \\
    \,\, + staircase & 19.1 & 24.40 \\
    \bottomrule
    \end{tabular}
    \caption{Performance of Arena-Hard and AlpacaEval2 (LC win rate). Results are shown for individual methods and combinations using the MoA framework with different aggregators.}
    \label{tab:diff_agg}
\end{table*}

\begin{table*}
    \centering
    \small
    \setlength{\tabcolsep}{2pt} 
    \begin{tabular}{@{}lccccc@{}}
    \toprule
    Method & Start Chunk Size & Second Chunk Size & Staircase Chunk Size & Staircase Agg Chunk Size & Arena-Hard \\
    \midrule
    First Chunk Ablat. & 4 & 128 & 256 & 128 & 47.9 \\
    First Chunk Ablat. & 32 & 128 & 256 & 128 & 48.5 \\
    Second Chunk Ablat. & 8 & 64 & 128 & 128 & 40.7 \\
    Second Chunk Ablat. & 8 & 256 & 512 & 128 & 42.6 \\
    Staircase Chunk Ablat. & 8 & 128 & 128 & 128 & 48.7 \\
    Staircase Chunk Ablat. & 8 & 128 & 512 & 128 & 48.7 \\
    \it{MoA staircase (reference)} & 8 & 128 & 256 & 128 & 47.5 \\
    \bottomrule
    \end{tabular}
    \caption{Results of ablation studies and reference configuration for Arena-Hard. Each row shows different configurations of chunk sizes and the corresponding performance on Arena-Hard. We specifically study the effect of varying chunk sizes. Abla. is abbreviation for ablation.}
    \label{tab:chunk_ablation}
\end{table*}

\begin{table*}[t]
    \centering
    \small
    \caption{Prompt template of prefix-caching optimized staircase streaming for MoA.}
    \begin{tabular}{@{}lp{0.9\linewidth}@{}}
    \toprule
    Role & Content \\
    \midrule
    system &
    You have been provided with a set of responses from various open-source models to the latest user query in chunks. Your task is to synthesize these response chunks into a single, high-quality response. It is crucial to critically evaluate the information provided in these responses, recognizing that some of it may be biased or incorrect.  As some responses may be incomplete yet, craft your synthesized response to allow for easy updating or expansion as new information becomes available. Your response should not simply replicate the given answers but should offer a refined, accurate, and comprehensive reply to the instruction. Ensure your response is well-structured, coherent, and adheres to the highest standards of accuracy and reliability. 
    \\ \\
    & Responses from models:
    \\
    & Chunk 1: \\
    & Model 1: \\
    & \{$R_{1,1}$\} \\
    & ... \\
    & Model $N$: \\
    & \{$R_{N,1}$\} \\
    \\
    & Chunk 2: \\
    & Model 1: \\
    & \{$R_{1,2}$\} \\
    & ... \\
    & Model $N$: \\
    & \{$R_{N,2}$\} \\
    \\
    & ... \\
    \midrule
    user & \{Q\} \\
    \bottomrule
    \end{tabular}
    \label{tab:template-moa-poss}
\end{table*}

\begin{table*}[t]
    \centering
    \small
    \caption{
        Prompt template of staircase streaming for MAD.
        MAD puts the model's own output before the debate prompt, here we assueme it's the `N`-th model.
    }
    \begin{tabular}{@{}lp{0.9\linewidth}@{}}
    \toprule
    Role & Content \\
    \midrule
    user & \{Q\} \\
    \midrule
    assistant & \{\{$\cup_{k=1}^{j} R_{N,k}$\}\} \\
    \midrule
    user &
    These are the responses to the query from other agents: \\
    \\
    & One agent solution: \{$\cup_{k=1}^{j} R_{1,k}$\} \\
    \\
    & One agent solution: \{$\cup_{k=1}^{j} R_{2,k}$\} \\ 
    \\
    & ... \\
    \\
    & One agent solution: \{$\cup_{k=1}^{j} R_{N-1,k}$\} \\
    \\
    & Using the responses from other agents as additional information, can you provide your response to the query?
The original query is \{Q\} \\
    \bottomrule
    \end{tabular}
    \label{tab:template-mad-ss}
\end{table*}



\section{Related Work}
\label{section:related_work_appendix}




\subsection{Multi-Agent Inference}

A straightforward solution to leverage multiple LLMs is ranking outputs from different models. 
\citet{jiang-etal-2023-llm} introduced \textsc{PairRanker}, which performs pairwise comparisons on candidate outputs to select the best one.
To alleivate the substantial computational costs of multi-LLM inference, other studies have explored training a \textit{router} that predicts the best-performing model from a fixed set of LLMs for a given input \citep{wang2024fusing,shnitzer2024large,lu2023routing}.
Additionally, FrugalGPT \citep{chen2023frugalgpt} proposed reducing the cost of using LLMs by employing different models in a cascading manner.
To better leverage the responses of multiple models, \citet{jiang-etal-2023-llm} trained \textsc{GenFuser}, 
a model designed to generate an improved response by capitalizing on the strengths of multiple candidates. 
\citet{huang2024enabling} proposed fusing the outputs of different models by averaging their output probability distributions.

Another line of work focuses on collaborative collaboration, where multiple LLMs act as agents that collectively discuss and reason through given problems interactively.
\citet{debate,mad,chan2023chateval} established a mechanism for discussions among LLM-based agents. 
ReConcile \citep{reconcile} adopt multi-agent discussion with weighted voting. 
\citet{wang2024rethinking} systematically compared collaborative approaches and found that a single agent with a strong prompt, including detailed demonstrations, can achieve comparable response quality to collaborative approaches. 
\citet{moe} adopted a layered structure, using different LLMs as proposers to propose answers and another LLM to aggregate the final answer.

\subsection{Efficient Inference}


Recent advancements in LLM inference optimization focus on enhancing system efficiency and resource utilization in order to reduce calculation. vLLM~\citep{vllm} addresses fragmentation with PagedAttention, which manages non-contiguous memory blocks efficiently. DeepSpeed-Inference~\citep{deepspeedinference} combines GPU, CPU, and NVMe memory for high-throughput inference of diverse transformer models. FlexGen~\citep{flexgen} increases the throughput with optimized GPU memory usage by integrating memory and computation from GPUs, CPUs, and disks, and quantizes weights to boost inference speed. Flash-Decoding~\citep{flashdecoding} based on FlashAttention~\citep{flashattn} accelerates long-context inference with parallel processing of KV pairs with length considered.

On the algorithm side, KV-Cache optimization is a commonly studied topic. During inference with LLM, it is necessary to store the KV pairs of previously generated tokens in a cache for future token generation.
As the length of generated tokens increases, this KV cache expands significantly, resulting in high memory consumption and longer inference times~\citep{kvquant,h2o,streamingllm}.
Another inference speedup technique is speculative decoding~\citep{speculative,speculativesampling}. 
It involves smaller models to have multiple educated token guesses in parallel, then validating and pruning these candidates based on their likelihood and consistency with the output, thus reducing computational load and improving performance. 
Following this idea, Staged Speculative~\citep{stagedspeculative} organizes speculative outputs into a tree structure, enhancing batch generation and overall performance. 
BiLD~\citep{bild} proposes a fallback policy allowing the small model to defer to the target model when uncertain, and a rollback policy for correcting small model errors.
\section{Effects of Different Aggregators}
We evaluate our approach to diverse aggregators to enhance our understanding of its generalizability.

In our initial experiments, we chose Gemma-2-9B-IT as the aggregator because the focus of the MoA architecture lies in optimizing the performance of the final aggregator. Through extensive testing, we observed that Gemma-2-9B-IT consistently outperformed other options and demonstrated superior robustness to staircase streaming. This motivated us to use it as the fixed aggregator for our primary experiments. We demonstrate the results of using different aggregators in \cref{tab:diff_agg} and we can see that our method is able to maintain the performances consistently. 

\section{Additional Ablations}
We conducted additional experiments to analyze the impact of varying chunk sizes. Using the configuration presented in \cref{tab:results} as a reference—specifically, \textit{Start Chunk Size = 8}, \textit{Second Chunk Size = 128}, and \textit{Staircase Chunk Size = 256}—we varied each parameter individually, testing both lower and higher values, and evaluated their effects on performance. Overall, as shown in \cref{tab:chunk_ablation}, we observed that performance remained fairly consistent across most configurations. However, setting the \textit{Second Chunk Size} to 64 resulted in a notable performance degradation. We hypothesize that this may be due to the importance of information in the early chunks, which the aggregator relies on for effective processing.

\begin{table*}[t]
    \centering
    \small
    \caption{Case study: staircase streaming of MoA for reasoning task. We pick an example from the MATH dataset. Different colors indicate different chunks. We omit some of the generations to fit into one page. Notice that even for reasoning tasks where important information (answer) is usually located at the end, our staircase setup can still synthesize the correct answer.}
    \begin{tabular}{@{}lp{0.6\linewidth}@{}}
    \toprule
    Role          & Content \\
    \midrule
    User          & Find the smallest integer that satisfies the inequality: [\textbackslash{}n(x-5)(x+5)<0.] \\
    \midrule
    Gemma-2-9B-IT          
    &  \sethlcolor{teal!60}\hl{Here's how to solve the inequality:}\sethlcolor{yellow!50}\hl{**1. Understand the Sign of the Product**The inequality $(x-5)(x+5) < 0$ tells us that the product of the two factors is negative.  This happens when one factor is positive and the other is negative.**2. Consider the Cases*** **Case 1:**  $x - 5 > 0$ and $x + 5 < 0$.  This leads to $x > 5$ and $x < -5$, which has no solutions since a number cannot be both greater than 5 and less than -5.* **Case 2:** }\sethlcolor{green!40}\hl{$x - 5 < 0$ and $x + 5 > 0$. This leads to $x < 5$ and $x > -5$.**3. Combine the Solutions**The solution to the inequality is $-5 < x < 5$.**4. Find the Smallest Integer**The smallest integer that satisfies this inequality is  -4.**Answer:** $\boxed{-4}$}

    \\   
    LLaMA-3.1-8B-Instruct            
    &  \sethlcolor{teal!60}\hl{To solve the inequality ((x-}\sethlcolor{yellow!50}\hl{5)(x+5)<0), we need to find the values of (x) for which the expression is less than zero.First, let's find the critical points by setting each factor equal to zero and solving for (x):1. $(x - 5 = 0 \implies x = 5)$2. $(x + 5 = 0 \implies x = -5)$These are the critical points where the expression changes sign. Now, we need to test the intervals between and beyond these points to see where the expression is negative.The intervals to test are:- }\sethlcolor{green!40}\hl{(x < -5)- (-5 < x < 5)- (x > 5)Let's test a value from each interval:- For (x < -5), let's test (x = -10): ((-10 - 5)(-10 + 5) = (-15)(-5) = 75 > 0)- For (-5 < x < 5), let's test (x = 0): ((0 - 5)(0 + 5) = (-5)(5) = -25 < 0)- For (x > 5), let's test (x = 10): ((10 - 5)(10 + 5) = (5)(15) = 75 > 0)From the tests, we see that the expression is negative only in the interval (-5 < x < 5).Therefore, the solution to the inequality is (-5 < x < 5), which can be written as (-5 < x < 5).}

    \\     
    Mistral-7B-Instruct-v0.3         
    & \sethlcolor{teal!60}\hl{ The smallest integer that satisfies the inequality }\sethlcolor{yellow!50}\hl{$(x-5)(x+5)<0$ is $x=4$.Here's the reasoning:1. The product of two numbers is less than zero if and only if one number is positive and the other is negative.2. The expression $(x-5)(x+5)$ will be positive when $x>5$ or $x<-5$ (since both factors are of the same sign), and negative when $-5<x<5$ (since one factor is positive and the other is negative).3. To find the smallest integer that satisfies the inequality, we look for the smallest integer within the inte}\sethlcolor{green!40}\hl{rval $-5<x<5$.4. Testing the integers in this interval, we find that $x=4$ is the smallest integer that makes the inequality true, since $(4-5)(4+5)=-9$ is less than zero.So, the final answer is $\boxed{4}$.}

    \\         
    Qwen-1.5-7B-Chat       
    & \dots

    \\ 
    \midrule
    \begin{tabular}{@{}l@{}}
    Aggregated \\
    (Gemma-2-9B-IT)
    \end{tabular}
  
    & \sethlcolor{teal!60}\hl{Here's how to find the smallest}\sethlcolor{yellow!50}\hl{ integer solution to the inequality $(x-5)(x+5) < 0$:**1. Understand the Sign of the Product**The inequality tells us that the product of the two factors is negative. This happens when one factor is positive and the other is negative.**2. Consider the Cases*** **Case 1:**  $x - 5 > 0$ and $x + 5 < 0$. This leads to $x > 5$ and $x < -5$, which has no solutions since a number cannot be both greater than 5 and less than -5.}\sethlcolor{green!40}\hl{* **Case 2:** $x - 5 < 0$ and $x + 5 > 0$. This leads to $x < 5$ and $x > -5$.**3. Combine the Solutions**The solution to the inequality is $-5 < x < 5$.**4. Find the Smallest Integer**The smallest integer that satisfies this inequality is -4.**Answer:** $\boxed{-4}$}

    \\ 
    \bottomrule
    \end{tabular}
    \label{tab:case}
\end{table*}

\begin{table*}[t]
    \centering
    \small
    \caption{Case study: staircase streaming of MoA for reasoning task. We pick an example from the MATH dataset. Different colors indicate different chunks. We omit some of the generations to fit into one page. Notice that even for reasoning tasks where important information (answer) is usually located at the end, our staircase setup can still synthesize the correct answer. This example contains more chunks.}
    \begin{tabular}{@{}llp{0.6\linewidth}@{}}
    \toprule
    Role        &  Preference   & Content \\
    \midrule
    User        &      -         & For how many different values of $x$ does $x^5 = x^4 + 72x^3$? \\
    \midrule
    Gemma-2-9B-IT            & 0.00 
    &  \sethlcolor{teal!60}\hl{Here's how to solve the problem:}\sethlcolor{yellow!50}\hl{**1. Rearrange the equation:**Subtract $x^4 + 72x^3$ from both sides to get a standard polynomial equation:$x^5 - x^4 - 72x^3 = 0$**2. Factor out a common factor:**Notice that all the terms have a common factor of $x^3$:$x^3(x^2 - x - 72) = 0$**3. Solve for x:**This equation has two parts:* **Part 1:**  $x^3 = 0$. This has one }\sethlcolor{green!40}\hl{solution: $x = 0$.* **Part 2:** $x^2 - x - 72 = 0$.  We can solve this quadratic equation using the quadratic formula:   $x = \frac{-b \pm \sqrt{b^2 - 4ac}}{2a}$   Where $a = 1$, $b = -1$, and $c = -72$.  Plugging in these values, we get:   $x = \frac{1 \pm \sqrt{(-1)^2 - 4 \cdot 1 \cdot -72}}{2 \cdot 1}$   $x = \frac{1 \pm \sqrt{289}}{2}$   $x = \frac{1 \pm 17}{2}$   This gives us two more solutions: $x = 9$ and $x = -8$.**4. Combine the solutions:**The total number of different solutions is 3: $x = 0$, $x = 9$, and $x = -8$.**Answer:** $\boxed{3}$}
       
    \\   
    LLaMA-3.1-8B-Instruct             & 0.00
    &  \sethlcolor{teal!60}\hl{To find the values of $x$ tha}\sethlcolor{yellow!50}\hl{t satisfy the equation $x^5 = x^4 + 72x^3$, we can rearrange the equation to get $x^5 - x^4 - 72x^3 = 0$. This is a polynomial equation of degree 5, and it can be factored by taking out a common factor of $x^3$ from all the terms, giving us $x^3(x^2 - x - 72) = 0$.Now, we have two factors: $x^3$ and $(x^2 - x - 72)$. The first factor, }\sethlcolor{green!40}\hl{$x^3$, equals zero when $x = 0$. For the second factor, $(x^2 - x - 72)$, we can factor it \dots is $9^5 = 59049$, but the right-hand side is}\sethlcolor{orange!30}\hl{$9^4 + 72(9)^3 = 6561 + 46656 = 52417$. This means that $x = 9$ is not a \dots The number of different values of $x$ that satisfy the equation is 2.$\boxed{2}$}
        
    \\     
    Mistral-7B-Instruct-v0.3         & 0.16
    & \sethlcolor{teal!60}\hl{ To find the values of $x$ tha}\sethlcolor{yellow!50}\hl{t satisfy the equation $x^5 = x^4 + 72x^3$, we can rearrange the equation to form a quadratic equation in $x^3$.First, subtract $x^4$ from both sides:$x^5 - x^4 = 72x^3$Next, factor out $x^3$ from the left side:$x^3(x^2 - x) = 72x^3$Now, divide both sides by $x^3$:$x^2 - x = 72$Finally, multiply both sides by $x$ to get a quadratic eq}\sethlcolor{green!40}\hl{uation in $x$:$x^3 - x^2 = 72x$This is a cubic equation in $x^3$, \dots we can express $x$ as a function of $x$:$x = \frac{3}{2} + \sqrt{72x + 81/4}$}\sethlcolor{orange!30}\hl{Square both sides to eliminate the square root:$x^2 = \frac{9}{4} + 72x + \frac{81}{16} + 2\sqrt{72x + 81/4}\cdot\sqrt{72x + 81/4}$Simplify the right side:$x^2 = \frac{9}{4} + 72x + \frac{81}{16} + 144\sqrt{(72x + 81/4)}$ \dots This is a quartic equation in $x^2$, which is di}\sethlcolor{red!20}\hl{fficult to solve analytically. However, \dots Now, square both sides again to eliminate the square root: }\sethlcolor{purple!20}\hl{\dots Unfortunately, it is not possible to find a single answer formula for the solutions of this equation. Instead, we can use numerical methods or a computer algebra system to find the real solutions. \dots}

    \\         
    Qwen-1.5-7B-Chat             & 0.03
    & \dots

    \\ 
    \midrule
    \begin{tabular}{@{}l@{}}
    Aggregated \\
    (Gemma-2-9B-IT)
    \end{tabular}
    & 0.33
    & \sethlcolor{teal!60}\hl{Here's how to solve the problem}\sethlcolor{yellow!50}\hl{:**1. Rearrange the equation:**Subtract $x^4$ and $72x^3$ from both sides to get a standard polynomial equation:$x^5 - x^4 - 72x^3 = 0$**2. Factor out a common factor:**Notice that all the terms have a common factor of $x^3$:$x^3(x^2 - x - 72) = 0$**3. Solve for x:**This equation has two parts:* **Part 1:**  $x^3$}\sethlcolor{green!40}\hl{ = 0. This has one solution: $x = 0$.* **Part 2:** $x^2 - x - 72 = 0$.  We can solve this quadratic equation using the quadratic formula:   $x = \frac{-b \pm \sqrt{b^2 - 4ac}}{2a}$   Where $a = 1$, $b = -1$, and $c = -72$.  Plugging in these values, we get:   $x = \frac{1 \pm \sqrt{(-1)^2 - 4 \cdot 1 \cdot -72}}{2 \cdot 1}$   $x = \frac{1 \pm \sqrt{289}}{2}$   $x = \frac{1 \pm 17}{2}$   This gives us two more solutions: $x = 9$ and $x = -8$.**4. Combine the solutions:**The total number of different solutions is 3: $x = 0$, $x = 9$, and $x = -8$.**Answer:** $\boxed{3}$}

    \\ 
    \bottomrule
    \end{tabular}
    \label{tab:case2}
\end{table*}

\end{document}